\documentclass[conference]{IEEEtran}
\IEEEoverridecommandlockouts
\usepackage{cite}
\usepackage{amsmath,amssymb,amsfonts}
\usepackage{algpseudocode}
\usepackage{graphicx}
\usepackage{textcomp}
\usepackage{xcolor}
\def\BibTeX{{\rm B\kern-.05em{\sc i\kern-.025em b}\kern-.08em
    T\kern-.1667em\lower.7ex\hbox{E}\kern-.125emX}}
\begin{document}

\title{Anomaly Detection in Power Markets and Systems\\

}

\author{\IEEEauthorblockN{Ugur Halden\IEEEauthorrefmark{1},
Umit Cali\IEEEauthorrefmark{1},
Ferhat Ozgur Catak\IEEEauthorrefmark{4}
Salvatore D'Arco\IEEEauthorrefmark{2}, 
Francisco Bilendo\IEEEauthorrefmark{3},}
\IEEEauthorblockA{\IEEEauthorrefmark{1}Norwegian University of Science and Technology, Høgskoleringen 1, 7491 Trondheim, Norway
}
\IEEEauthorblockA{
\IEEEauthorrefmark{2}Sintef Energy Research, Sem Sælands vei 11, 7034 Trondheim, Norway}
\IEEEauthorblockA{
\IEEEauthorrefmark{3}Nanjing University of Aeronautics and Astronautics, Nanjing, China}
\IEEEauthorblockA{
\IEEEauthorrefmark{4}University of Stavanger, Kjell Arholms gate 41, 4021 Stavanger, Norway}

}

\maketitle

\begin{abstract}
The widespread use of information and communication technology (ICT) over the course of the last decades has been a primary catalyst behind the digitalization of power systems. Meanwhile, as the utilization rate of the Internet of Things (IoT) continues to rise along with recent advancements in ICT, the need for secure and computationally efficient monitoring of critical infrastructures like the electrical grid and the agents that participate in it is growing. A cyber-physical system, such as the electrical grid, may experience anomalies for a number of different reasons. These may include physical defects, mistakes in measurement and communication, cyberattacks, and other similar occurrences. The goal of this study is to emphasize what the most common incidents are with power systems and to give an overview and classification of the most common ways to find problems, starting with the consumer/prosumer end working up to the primary power producers. In addition, this article aimed to discuss the methods and techniques, such as artificial intelligence (AI) that are used to identify anomalies in the power systems and markets.
\end{abstract}


\section{Introduction}
A stable and efficient electrical infrastructure has always been challenging to operate and maintain \cite{9837910}. The power networks have significantly increased in complexity during the past century as a result of the rising demand for electricity. Due to the increased percentage of Renewable Energy Sources (RES) that are intermittent and variable, the future power system will have to manage an ever-increasing level of complexity. The expansion of RES, which are implemented in order to meet the climate targets of 2050 \cite{nikas2021eu}, will be made possible by ensuring a reliable electricity infrastructure. With the advent of RES, the grid is now influenced not only by the fluctuating levels of demands but also by the fluctuating levels of electricity generation. In order to maintain the reliability and availability, this complexity necessitates accurate measurement and analysis techniques.

Traditional methods for power grid State Estimation (SE) collect data from connected Remote Terminal Units (RTUs) installed in the grid and monitor it using Supervisory Control and Data Acquisition (SCADA) systems \cite{8624411}. These RTUs gather information on the voltage levels, as well as the active and reactive power sources nearby. The technique gives a summary of the grid and a SE because the system gathers data as scans over a period of time. However, if a change or an abnormal event happened during or after the time interval, the measurements may not be applicable to the state of the power grid due to the fact that the RTUs are not time synchronized \cite{karpilow}.

Phasor Measurement Units (PMUs) have raised the possibility of more frequent grid measurements, which in return has raised the potential for grid SE as well. A more precise and real-time image of the condition of the power grid is possible due to the time-synchronized measurements. Using a measuring frequency that is equal to or close to the grid frequency, the PMUs can deliver real-time online measurements across the grid. This makes timely anomaly identification and continuous SE possible. The disadvantages, nevertheless, are related to the volume of data. Millions of data points are swiftly accumulated with a high measurement frequency, where each PMU may record voltage magnitude, angle, frequency, and current. Such massive volumes of data demand specialized techniques and procedures for processing and analysis \cite{regev2022hybrid}.

\begin{figure*}[t]
    \centering
    \includegraphics[width=0.80\textwidth]{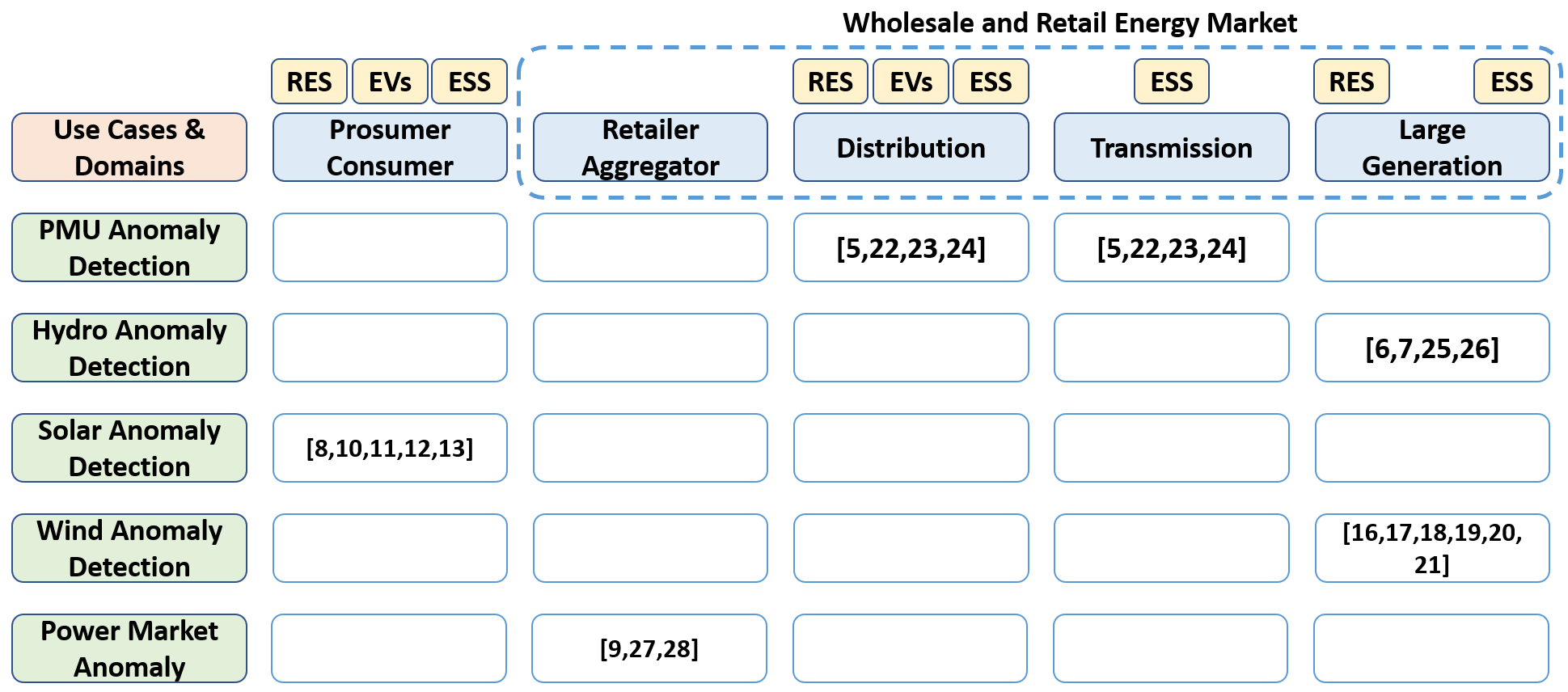}
    \caption{Overview and segmentation of anomaly detection in power markets and systems.}
    \label{fig:anomaly_segmentation}
\end{figure*}

As can be seen from Figure \ref{fig:anomaly_segmentation}, PMUs largely operate in transmission and distribution domains. Mainly due to their associated high costs and lower need to perform SE in the residential domain. However, with the rise of Distributed Energy Resources (DERs), faults and anomalies can also happen in the consumer/prosumer side and in the large scale generation side due to intermittency of RES. Additionally, as systems become increasingly interconnected, cyberattacks have the potential to disrupt every domain and can lead to catastrophic consequences.

\section{Types of anomalies and their consequences on power systems and markets}
Anomalies are inevitable when operating a complex system such as power network, and given its complexity, many distinct types of anomalies may occur. This section will provide information on various power system anomalies.

\subsection{Bad Data and False Data Injections}
While measuring and transferring large number of data points, it is probable that some errors will occasionally occur \cite{karpilow}. A brief burst of inaccuracy during data transmission might cause a significant amount of missed data points, especially in the case of PMUs. However, as PMU measurements are time synchronized, the missing data can proactively be included in the end dataset. This will nevertheless cause issues for the system's real-time SE or anomaly detection, which depend on a constant data flow.

False Data Injection (FDI) based attacks operate by gaining unrestricted access to the data channel and changing the measurements that are retrieved between the measurement and data retrieval points. FDI-style attacks can have a significant impact on the power grid's ability to operate steadily because measured data is constantly used for load balancing, SEs, and production. Hence, FDIs have the potential to quickly destabilize the entire system, with potentially disastrous results.

\subsection{Physical Component Errors}
The modern power systems are composed of various components such as small and distributed PV panels in the prosumer end to large scale generators such as wind and hydro. Thus, as the each unit nears its life cycle, physical errors will occur. In the case of hydro, one example can be various vibration related errors in the generator, which will need downtime for required maintenance \cite{Hoh2022AData}. In the case of wind energy, gearbox wear can result in dangerous operation of the wind turbines and will need emergency maintenance \cite{Konig2021MachineSystems}. When it comes to PV systems, Short Circuit (SC) or Open Circuit (OC) errors can arise as well as various other inverter related errors and physical errors at the PV panels themselves \cite{HAJJI2021313}.

\subsection{Market Manipulation}
As electricity is being regarded and traded as a commodity, various manipulations can occur in the wholesale market in order to give unethical advantage towards one or more participating agent. This can include; physical withholding of energy or wash trades to artificially drive up the prices, or pump-dump, pre-arranged trading to exclude healthy competition from the market \cite{Gren2019AnAnalysis}. Meanwhile, transmission capacity hoarding without the intent of use can lead to trade interrupts and depending on number and location of pricing regions, can lead prices to increase artificially, damaging the end users and other grid agents.

\section{AI/ML/Statistical Methods to Detect Anomalies}
\subsubsection{One-Class Support Vector Machine (OCSVM)}

Support Vector Machine (SVM) is a supervised learning algorithm to solve classification and regression problems. SVM works by finding a hyperplane that separates the classes from each other. In one-class SVM, the model is trained on a single class to find a hyperplane that separates the data from the rest of the data. The hyperplane is found by maximizing the distance between the hyperplane surface and the data. The hyperplane that maximizes the distance is the one that is most similar to the training data. OCSVM is defined as follows:
\begin{equation}
    \min_{w, \xi, \rho} \frac{1}{2}w^Tw + \frac{\nu}{N}\sum_{i = 1}^N \xi_i + \frac{1}{2}\sum_{i = 1}^N \sum_{j = 1}^N y_i y_j K(x_i - x_j) \xi_i \xi_j
\end{equation}
$\nu$ is a parameter that determines the trade-off between the distance from the hyperplane with the number of points that are on the wrong side of the hyperplane. $K$ is a kernel function, $N$ is the number of training data points, and $\xi_i$ is the slack variable. The slack variable is used to allow for training data that is on the wrong side of the hyperplane. Training data on the wrong side of the hyperplane is considered an outlier. SVM can be used for anomaly detection by mapping the data to a higher dimensional space and using a hyperplane to decide the boundary between the inliers and outliers.

\subsubsection{Long Short Term Memory (LSTM)}
A common and effective deep neural network approach for time series data analysis is Long Short-Term Memory (LSTM) which is made up of an input layer, one or more hidden layers, memory cells, and an output layer. By utilizing forget and memory cells, the algorithm can preserve temporal relations in the dataset. 

\begin{equation}
    \mathbf{I_t} = \sigma\left(\mathbf{X}_t\mathbf{W}_{xi} + \textbf{H}_{t-1}\mathbf{W}_{hi}+\mathbf{b}_i\right)
    \label{Eq:It}
\end{equation}
\begin{equation}
    \mathbf{F_t} = \sigma\left(\mathbf{X}_t\mathbf{W}_{xf} + \textbf{H}_{t-1}\mathbf{W}_{hf}+\mathbf{b}_f\right)
    \label{Eq:Ft}
\end{equation}
\begin{equation}
    \mathbf{O_t} = \sigma\left(\mathbf{X}_t\mathbf{W}_{xo} + \textbf{H}_{t-1}\mathbf{W}_{ho}+\mathbf{b}_o\right)
    \label{Eq:Ot}
\end{equation}

The computation of each gate's output value is shown in the equations \ref{Eq:It}, \ref{Eq:Ft} and \ref{Eq:Ot} where $\mathbf{F_t}$ represents the forget gate, $\mathbf{I_t}$ input and $\mathbf{O_t}$ output gates. The activation function is denoted by $\sigma$ in this case, while the weight parameters are denoted by $\mathbf{W}$ and the bias parameters by $\mathbf{b}$. The input value to the layer and the output of the hidden layer from the previous step are $\mathbf{X_t}$ and $\mathbf{H}_{t-1}$ respectively.

\subsubsection{Kernel density estimation (KDE)}

Kernel density estimation (KDE) is an unsupervised learning algorithm used to estimate a dataset's Probability Density Function (PDF). The algorithm uses a set of kernels to approximate the underlying density function of a dataset. A kernel function is a function that returns a value between 0 and 1. The sum of all of the kernels is equal to 1. The probability of a data point x equals the sum of the kernel functions at x. The kernel density estimation algorithm is defined as follows:
\begin{equation}
    f_h(x) = \frac{1}{nh}\sum_{i = 1}^n K(\frac{x - x_i}{h})
\end{equation}
where $n$ is the number of training data points, $x_i$ is the i-th training data point, and $K$ is the kernel function. The parameter h is the bandwidth. Anomaly detection with kernel density estimation is done by looking at the value of the kernel density function at the test data point, if below a threshold, it can be classified as an anomaly. 




\subsubsection{Autoencoders (AE)}

Autoencoders are unsupervised learning algorithms used to learn low-dimensional data representations and are composed of an encoder and a decoder. The encoder maps the input to a lower dimensional space, and the decoder maps the lower dimensional space back to the original dimension. The encoder and decoder are trained with the goal of reconstructing the original input. The training dataset is used to train the encoder and decoder, and the loss is calculated as the difference between the original input and the reconstructed input. Anomaly detection with an autoencoder is done by training an autoencoder on the training dataset and checking the loss of the test dataset. A high loss means that the reconstructed input is not similar to the original input, which means that the data point is an anomaly. Autoencoder-based anomaly detection is defined as follows
\begin{equation}
    \min_{W, b} \frac{1}{N}\sum_{i = 1}^N ||x_i - \hat{x}_i||^2
\end{equation}
where $W$ and $b$ are the weights and biases, $x_i$ is the original input, and $\hat{x}_i$ is the reconstructed input. If the distance between the original input, $x_i$, and reconstructed input,  $\hat{x}_i$, is larger than a threshold value, then $x_i$ is classified as an anomaly.

\section{Systematic Overview of Anomaly Detection Use Cases on the Power Market and Systems Domains}

As the deployment of PV systems continues to rise, it is essential to ensure the system operates in a stable and safe manner. However, faults continue to be a challenge for the effective operation of PV systems despite technological improvement and the adoption of the newest standards. As far as possible, a holistic approach should be employed to detect and diagnose at least all of the most common and serious errors in a PV system while using ML to analyze the fault \cite{tarikua_pv}.

In \cite{HAJJI2021313}, authors developed a feature extraction stage based on principal component analysis and tested various ML based classifiers such as k-Nearest Neighbour (k-NN), Random Forest (RF), Naïve Bayes (NB), SVM, etc. for detection of multiple faults in a modelled 15 kWp system (OC, sensor faults, shading). The results of the study indicates the models all achieved accuracies greater than 96\%. 

Meanwhile, Li et al. \cite{s20174688} proposes comparison between the real and modelled system outputs for fault detection. Various ML models such as Multilayer Perceptron (MLP), k-NN, SVM were utilized for detection of faults such as OC, SC, shading. According to the published results, the best accuracy was achieved with MLP. Additionally in \cite{Basnet2020AnSystems}, the authors also proposes utilization of MLP to identify and detect OC and SC faults where the model is able to achieve 96\% accuracy. In addition, Artificial Neural Networks can also be utilized for fault detection in PV systems as performed by \cite{8895279} and can achieve accuracies greater than 99\%.

With regards to anomaly detection, wind turbines condition monitoring approaches can be divided into two different categories: (i) standalone condition monitoring system (CMS)-based, and (ii) SCADA system-based. For instance, standalone CMS-based methods include vibration analysis, lubrication analysis, acoustic emission analysis and electrical analysis \cite{wind_1, wind_2}. 

Although the aforementioned standalone CMS-based methods have attained substantial results in terms of diagnosis, the main drawback includes the fact that they are commonly found to be expensive since they require additional equipment investment, calibration, and installation \cite{7084135}. Alternatively, there is already a substantial amount of information in large utility scale wind turbines, which is mainly used at the level of safeguarding, originating from the built-in SCADA system. Making use of this information already available at no additional costs is a cost-efficient alternative to improving the reliability of wind turbines and reducing the O\&M costs of wind farms.

It is however important to mention that, wind turbine SCADA data are generally recorded with an interval of 10 min average per sample point; they are typically a collection of statistical features (e.g., mean, maximum and minimum values, and standard deviation) of a wide range of signals (e.g., temperature, current, voltage, active power, rotor speed, wind speed, etc.). Since the signals are recorded with a long interval, most dynamic features of wind turbine anomalies and faults which are useful for condition monitoring may be lost; consequently, the detailed information, for instance the location and mode of most wind turbine anomalies and faults, cannot be diagnosed by using the conventional signal processing techniques such as time domain, frequency domain, and time-frequency domain analyses \cite{ZENG2020106233}. Ideally, methods based on SCADA data usually resort to purely statistical or ML methods. Examples of such methods can be found in \cite{wind_3, MEYER2021117342, wind_4, RENSTROM2020647} where the authors explored methods in the framework of ML such as ensemble model, normal behavior model plus automatic techniques such as change-point detection, and deep learning model such as autoencoder, among others.

While operating a vastly interconnected and complex system such as power grid, various anomalies can occur such as bad data due to measurement, communication errors or cyber attacks, topological errors, impulsive and oscillatory transients, sudden load changes, etc. Due to major economical costs associated with power system errors, it is important to increase the dependability of power systems and markets \cite{regev2022hybrid}. 

An Artificial Neural Network (ANN) based on the Group Method of Data Handling (GMDH) is used by the authors of \cite{yuval_1} to determine where the error has occurred and to distinguish between topological and analogical faults. The proposed methodology was tested in a Brazilian 118-bus system and shows promising results in real time anomaly detection and able to distinguish between topological and analogical errors.

Meanwhile, authors in \cite{RAMANMR2020100393} proposed MLP and cumulative sum technique to detect stealth based attacks on cyber-physical systems such as power grid. Utilized MLP and cumulative summing technique was able to detect FDIs even if they were discrete or continuous. Similarly, \cite{regev2022hybrid} also focused on FDI attacks on power systems as PMU datasets are especially vulnerable against them. For this research, the authors compared various ML algorithms such as Convolutional Neural Network (CNN), LSTM and their hybridized versions. According to the results the most prominent accuracies were retrieved from C-LSTM with 96.89\%. 

In addition, purely statistical methods can also be utilized for anomaly detection in electrical grids as performed by \cite{yuval_2} where number  of different methods were utilized such as Discrete Fourier Transform (DFT), Discrete Wavelength Transform (DWT), shapelet derivative, etc. The results indicate shapelet derivative outperforms other techniques and able to rapidly detect and classify anomalies in power system. 

Similar to solar and wind energy, development of environmental hydro power is also essential for achieving decarbonization of the energy sector.  Thus, like wind power, increasing the working time between scheduled maintenances, and detecting any possible faults as soon as possible is an important aspect in hydro power. 

The authors in \cite{Bascones} utilize a method based on creation of behavioral patterns by k-Means and Self Organizing Maps (SOMs) which is later fed into Probability Density Distributions (PDDs). According to the results, the hybridization of both algorithms is able to detect anomalies happening in bearing temperature of hydro power generators and even able to correctly identify scheduled maintenances, disregarding them as faults. Comparatively, Hoh et al. \cite{Hoh2022AData} proposes a Generative Adversial Network (GNN) based technique for anomaly detection in hydro pumps. According to the results, the system is capable of working under high frequency data input and can classify an input data as normal or anomalous by comparing and assigning an anomaly score to the real and reconstructed input signals.

In \cite{Konig2021MachineSystems}, the authors were able to accuratelly detect and classify wear behavior of sliding bearings into three categories such as; running in, bad lubrication and contaminated oil. The developed methodology is based on CNNs and acoustic emissions. A simmilar three categorization was also performed by \cite{Quatrini2020MachineActivities} where faults were classified into: "Expected, Warning and Critical". For this classification, the authors compare both Random Forest Algorithm (RFA) and Decision Jungle Algorithm (DJA) where the latter underperforms compared to RFAs by a factor of 8.4\%. 

In \cite{Gren2019AnAnalysis} the authors are focusing on anomaly detection in power markets. First the historical data is split between historical and one week before the monitoring. Later, both data are compared. If an outlier has been found, a further regression analysis while considering external inputs such as weather and transmission capacity is performed. If the predicted bidding values still looks anomalous compared to actual values, the algorithm flags bids as potential market violation so a human agent in the loop can have a further in-depth look. 

Meanwhile, Sun et al. \cite{9482640} proposed a methodology based on inter-working three components as; LSTM and probabilistic local price forecasting coupled with probabilistic anomaly detection. The results indicate F1-Scores between 0.65 to 0.89 and can detect various cyber attacks towards the energy market such as Load Redistribution Attack (LRA) and Price Responsive Attack (PRA). Similarly, the researchers in \cite{9351299} also focused on market manipulation and compared 2 anomaly detection algorithms based on CNNs and SVMs. According to the results the CNN outperforms SVMs and can detect malicious price variations between 4-16\%.

\section{Conclusions and Outlook}

Operating and maintaining a reliable and efficient electrical infrastructure has always been a challenge. This article sought to investigate and discuss the methodologies and strategies used to detect abnormalities in power systems and markets, using various methods such as AI-based advanced analytics. PMU provide real-time online measurements throughout the grid, allowing for quick anomaly discovery. FDI attacks work by getting unfettered access to the data channel and altering the measurements collected between measurement and data retrieval points. By translating the data to a higher dimensional space and utilizing a hyperplane to determine the border between inliers and outliers, SVM may be used to identify anomalies. LSTM and other successful data analytics methods are used quite often in this field. SCADA data provides the most dynamic characteristics of wind turbine anomalies and faults, making it helpful for condition monitoring. 

The value of resilience in electricity systems and markets is growing, since the social and economic consequences are getting more important. Recent advancements in enabling technologies such as ICT, IoT, AI, and ML should be evaluated on a regular basis to maintain the steady and efficient functioning of key infrastructure and also anomaly detection purposes to increase the resilience capabilities of the modern power systems.

The recent developments in literature, particularly in enabler technologies such as ICT, IoT, AI and ML should continuously be assessed in order to ensure stable and efficient operation of critical infrastructure such as power grid. With the recent recent threats regarding the future of distributed power systems and market's cyber-physical security, the importance of resilience in power grids and markets are becoming more invaluable where the social, political and economical ramifications can be intolerable.

\bibliographystyle{IEEEtran}
\bibliography{references}
\end{document}